\documentclass[10pt,twocolumn,letterpaper]{article}

\usepackage[pagenumbers]{cvpr} % To force page numbers, e.g. for an arXiv version

\usepackage{graphicx}
\usepackage{amsmath}
\usepackage{amssymb}
\usepackage{booktabs}

\usepackage{multirow}
\usepackage{verbatim}

\usepackage[pagebackref,breaklinks,colorlinks]{hyperref}

\usepackage[capitalize]{cleveref}
\crefname{section}{Sec.}{Secs.}
\Crefname{section}{Section}{Sections}
\Crefname{table}{Table}{Tables}
\crefname{table}{Tab.}{Tabs.}

\def\cvprPaperID{1396} % *** Enter the CVPR Paper ID here
\def\confName{CVPR}
\def\confYear{2023}

\newcommand{\aaa}[1]{\textcolor{red}{\small (#1)}}
\newcommand{\bbb}[1]{\textcolor{blue}{\small (#1)}}

\begin{document}

\title{Prompt-Matched Semantic Segmentation}

\author{Lingbo Liu$^1$~~~ Jianlong Chang$^2$~~~ Bruce X.B. Yu$^1$~~~ Liang Lin$^3$~~~ Qi Tian$^2$~~~ Chang-Wen Chen$^1$ \\
$^1$The Hong Kong Polytechnic University, Hong Kong \\
$^2$Huawei Cloud \& AI, Shenzhen 518000, China~~~~  $^3$Sun Yat-Sen University, China
}

\maketitle

%%%%%%%%% ABSTRACT
\begin{abstract}
The objective of this work is to explore how to effectively and efficiently adapt pre-trained visual foundation models to various downstream tasks of semantic segmentation. Previous methods usually fine-tuned the entire networks for each specific dataset, which will be burdensome to store massive parameters of these networks. A few recent works attempted to insert some extra trainable parameters into the frozen networks to learn visual prompts for parameter-efficient tuning. However, these works showed poor generality as they were designed specifically for Transformers. Moreover, using limited information in these schemes, they exhibited a poor capacity to learn beneficial prompts.
To alleviate these issues, we propose a novel Stage-wise Prompt-Matched Framework for generic and effective visual prompt tuning. Specifically, to ensure generality, we divide the pre-trained backbone with frozen parameters into multiple stages and perform prompt learning between different stages, which makes the proposed scheme applicable to various architectures of CNN and Transformer. For effective tuning, a lightweight Semantic-aware Prompt Matcher (SPM) is designed to progressively learn reasonable prompts with a recurrent mechanism, guided by the rich information of interim semantic maps. Working as deep matched filter of representation learning, the proposed SPM can well transform the output of the previous stage into a desirable input for the next stage, thus achieving the better matching/stimulating for the pre-trained knowledge.
Extensive experiments on four benchmarks demonstrate that the proposed scheme can achieve a promising trade-off between parameter efficiency and performance effectiveness. Our code and models will be released.
\end{abstract}

%%%%%%%%% BODY TEXT
\section{Introduction}
Semantic segmentation \cite{minaee2021image} is a fundamental yet challenging visual problem in computer vision, which aims to automatically perform pixel-level labeling with a set of object categories for the given image. Over the past decades, this problem has attracted extensive research in industry and academia, since it can facilitate a variety of downstream applications, e.g., scene understanding \cite{long2015fully}, satellite image analysis \cite{liu2022aerial}, and medical auxiliary diagnosis \cite{he2019non}.

In the literature, various Convolutional Neural Networks (CNN) and Transformer architectures have been proposed for this problem \cite{badrinarayanan2017segnet,liu2018path,yu2018learning,ding2019semantic,li2020semantic,huang2021alignseg,strudel2021segmenter,xie2021segformer,gu2022multi}. Instead of learning from scratch \cite{zhu2019scratchdet}, most previous works usually first acquired a foundation model pre-trained on large-scale benchmarks (e.g., ImageNet \cite{deng2009imagenet}), and then fine-tuned the network's parameters on specific downstream datasets. There are usually two strategies for model fine-tuning. The first one is {\bf\textit{Full-Tuning}} \cite{agrawal2014analyzing} that adjusts all parameters of the whole network. However, it usually requires large amounts of training data with rich annotations for effective representation learning. Moreover, this strategy requires storing a proprietary model with massive parameters for each downstream task/dataset, which is burdensome and unsustainable for many service platforms. The second strategy is {\bf\textit{Head-Tuning}} \cite{donahue2014decaf}, which freezes the parameters of the backbone network and only optimizes the model's head. Intuitively, all tasks share the same backbone and we only need to maintain an individual head for each task. Despite being parameter efficient, this strategy has a limited capacity to learn discriminative features and can not well exploit pre-trained knowledge for complex visual understanding. Overall, the above strategies suffer from various issues, and we desire a more effective and efficient fine-tuning method for widespread downstream tasks of semantic segmentation.

Recently, {\bf\textit{Prompt-Tuning}} \cite{liu2021pre} has been proposed to explore the knowledge of frozen language models by inserting textual prompts into the downstream input. Inspired by its great success in natural language processing (NLP), a few works \cite{jia2022visual,chen2022adaptformer} in the field of computer vision have attempted to introduce some trainable prompt parameters to energize the pre-trained knowledge in those parameter-frozen visual foundation models. Despite certain progress, these works suffered from the following issues.
\textbf{First}, they were specially designed for Transformer, not generic to commonly-used CNN architectures. For instance, VPT \cite{jia2022visual} learned visual prompts in the token space, while AdaptFormer \cite{chen2022adaptformer} replaced the original MLP block with a parallel bottleneck module.
\textbf{Second}, these works modified the original structure of every foundation unit. In practice, the different stages of foundation models with standard structures are usually compressed and embedded into high-speed inference devices \cite{ma2017end,wang202228nm}. However, extensive structural modifications make the above approaches not directly applicable on these devices.
\textbf{Third}, these works learned visual prompts with limited information in a black-box mapping manner. Specifically, they only utilized the final recognition loss to optimize prompt parameters and had limited capacities to learn reasonable prompts. It is worth noting that image semantic segmentation is more challenging than image recognition, requiring richer information to perform pixel-wise inferences. Therefore, informative knowledge should be fully explored to generate beneficial visual prompts for parameter-efficient representation learning.

To address the above issues, we propose a novel Stage-wise Prompt-Matched Framework, which can effectively adapt those frozen foundation models of different architectures to facilitate semantic segmentation in various scenarios. Specifically, instead of modifying each foundation module with extra parameters, we partition the given backbone network into multiple stages and perform stage-wise prompt tuning for specific datasets. This makes our method applicable to various backbone architectures and those stage-wise embedded high-speed devices.
Moreover, for effective and efficient tuning, we introduce a lightweight trainable module termed Semantic-aware Prompt Matcher (SPM) that incorporates rich information of interim semantic maps to learn reasonable visual prompts between any two stages in a progressive and recurrent manner. Working like deep matched filtering, our SPM can effectively transform the output representation of the previous stage into an appropriate input representation for the next stage, making it better to match/stimulate the pre-trained knowledge in the frozen backbone.
Finally, we apply the proposed SPM to fine-tune various backbone networks such as ResNet \cite{he2016deep} and Vision Transformer \cite{dosovitskiy2020image} to handle semantic segmentation of natural, satellite, and medical images. Extensive experiments conducted on four benchmarks show the parameter efficiency and performance effectiveness of the proposed method. In particular, optimizing only a small number of parameters of our SPM and the head segmenter, our method not only significantly outperforms existing parameter-efficient fine-tuning methods, but also is comparable or even better than \textit{Full-Tuning}.

In summary, the contributions of our work are three-fold:
{
\begin{itemize}
\setlength{\itemsep}{0pt}
\setlength{\parsep}{0pt}
\setlength{\parskip}{1pt}
\item A novel Stage-wise Prompt-Matched Framework is proposed to effectively and efficiently fine-tune those pre-trained foundation backbones with frozen parameters. Without specifically modifying the structure of foundation units, our method is universal for various network architectures of CNN and Transformer.
\item A lightweight SPM is introduced to progressively learn reasonable visual prompts between different stages of the backbone through a recurrent mechanism. Guided by the rich information of interim semantic maps, our SPM can well transform the output of the previous stage into an appropriate input for the next stage, thus better energizing the pre-trained knowledge.
\item Extensive experiments on four benchmarks demonstrate that our method is performance-effective and parameter-efficient for fine-tuning foundation models to various downstream tasks of semantic segmentation.
\end{itemize}
}

\begin{figure*}[t]
  \begin{center}
     \includegraphics[width=2.1\columnwidth]{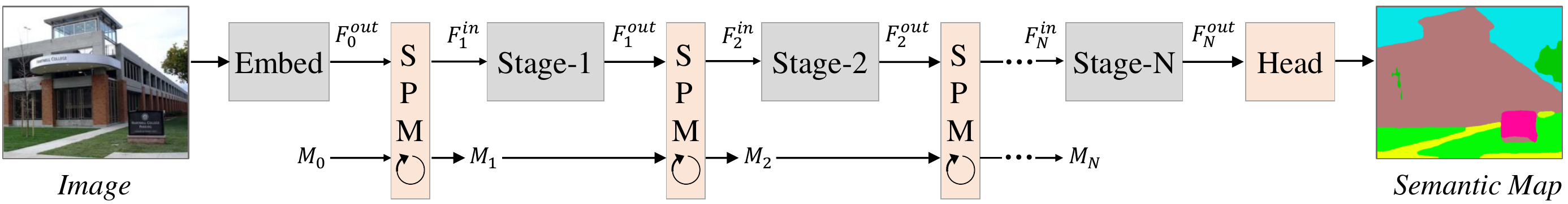}
  \vspace{-10mm}
  \end{center}
   \caption{The architecture of our Stage-wise Prompt-Matched Framework for semantic segmentation prompt tuning. A lightweight Semantic-aware Prompt Matcher (SPM) is introduced to learn reasonable visual prompts recurrently between every two stages of the frozen backbone network. Guided by the rich information of interim semantic maps, the $i$-th SPM transforms the output feature $F_{i-1}^{out}$ of the previous stage into a suitable input feature $F_{i}^{in}$ for the next stage. $M_i$ is the interim semantic map generated at the $i$-the SPM, while the initial map $M_0$ is generated from the statistic category probability of the training set. The symbol $\circlearrowright$ denotes the recurrent prompt learning mechanism in our SPM. It is worth noting that only the parameters of our SPM and head segmenter are updated during the training phase.}
\vspace{-1mm}
\label{fig:network-structure}
\end{figure*}

\section{Related Work}
{\bf{Prompt Tuning: }}
In recent years, various large-scale NLP models such as BERT \cite{devlin2018bert}, GPT-3 \cite{brown2020language}, and Pangu-$\alpha$ \cite{zeng2021pangu} have been developed by pre-training on huge datasets. With the emerging prompt tuning, these large-scale models achieved impressive transfer performance on myriads of downstream tasks such as translation \cite{tan2022msp}, reading comprehension \cite{hu2022knowledgeable}, question answering \cite{yang2022empirical}, etc. Inspired by the NLP prompt tuning paradigm, some computer vision researchers have attempted to facilitate visual understanding by fine-tuning vision-language models, where the textual prompts generated by text encoders were used to guide the representation learning of visual encoders \cite{radford2021learning,jia2021scaling}. Despite achieving promising performance, these works relied heavily on the textual prompt design and can not be smoothly applied to various vision tasks. With this concern, a few recent works \cite{jia2022visual,chen2022adaptformer} introduced some extra trainable parameters to directly learn visual prompts from the input visual features or randomized noise. Nevertheless, they were specially designed for Transforms and may fail to generate effective prompts due to the scarcity of instructive information. Therefore in this work, we are committed to exploring general and effective strategies for visual prompt tuning.

{\bf{Semantic Segmentation: }}
%Semantic segmentation is a fundamental yet challenging problem in computer vision.
As a typical pixel-wise prediction problem, semantic segmentation has been significantly promoted by deep neural networks \cite{lin2018multi,huang2019ccnet,zhong2020squeeze,hu2021region,li2022deep,chen2022vision}. For instance, \cite{long2015fully} proposed Fully Convolutional Networks that replaced fully-connected layers with convolutional layers to handle images of arbitrary sizes. \cite{ronneberger2015u} applied a convolutional encoder-decoder architecture with skip connections to generate semantic maps with high resolutions. \cite{zhao2017pyramid} used a backbone network to extract the feature maps of input images and then introduced a pyramid pooling module to aggregate different sub-region representations for multiscale contextual modeling. DeepLab family \cite{chen2015semantic,chen2017deeplab} applied dilated convolutions to enlarge the network receptive fields and introduced Conditional Random Fields to refine the final segmentation results. Recently, Transformers \cite{vaswani2017attention} have also been applied to address this problem. One representative work is Segmenter \cite{strudel2021segmenter}, which divided the image into local patches and fed their linear embeddings into Vision Transformer \cite{dosovitskiy2020image} to capture global context at each layer for semantic segmentation.
Despite progress, previous methods usually fine-tuned the parameters of the whole networks respectively for each specific task of semantic segmentation. It is burdensome to store massive parameters of these models, especially on some resource-constrained devices. Under this circumstance, we crave for a novel fine-tuning approach, where these tasks can share most of the parameters and achieve competitive results.

\section{Methodology}
\subsection{Overall Architecture}
In this work, we aim to utilize visual prompt learning to fine-tune the pre-trained foundation models for various downstream tasks, e.g., semantic segmentation. When developing our algorithm, we consider the following two questions: {{i)}} {\bf\textit{where to learn visual prompts}} and {{ii)}} {\bf\textit{how to learn reasonable prompts}}? We argue that previous works \cite{jia2022visual,chen2022adaptformer} that performed customized prompt learning within each foundation unit are not applicable to different network architectures. We also observe that it is suboptimal to learn prompts with limited information of the final loss of the model's head. So we desire an architecture-general prompt tuning model that fully exploits rich information to learn effective visual prompts.

To this end, we propose a unified Stage-wise Prompt-Matched Framework to effectively and efficiently fine-tune pre-trained foundation models to deal with downstream visual tasks. Here we take semantic segmentation as an example to illustrate the working process of our method. As shown in Figure \ref{fig:network-structure}, a semantic segmentation model usually consists of a universal backbone network pre-trained on a large-scale dataset and a customized head segmenter with random initialization. To reduce the number of tunable/stored parameters, we freeze the backbone network so that it can be shared by different tasks of semantic segmentation, while the head segmenter is optimized for each specific dataset. Inspired by previous NLP works \cite{liu2021pre}, we apply prompt tuning to efficiently exploit the backbone pre-trained knowledge to facilitate the visual representation learning of downstream semantic segmentation. To make our method applicable to various network architectures, we propose to learn visual prompts between different stages of the frozen backbone, without modifying the original structures of foundation units, e.g., residual module or transformer layer. Specifically, based on its architecture, we partition the backbone network into $N$ stages, each of which is composed of multiple foundation units. Notice that there may be some embedding layers before the first stage. For convenience, the output feature of stage $i$ is denoted as $F_i^{out}$ ($i$=1,...,$N$), while the output feature of those embedding layers is denoted as $F_0^{out}$.

We then introduce a differentiable Semantic-aware Prompt Matcher (SPM) to learn visual prompts between two adjacent stages using a small number of parameters. As mentioned above, rich information is desired to perform prompt learning for various vision tasks including semantic segmentation. In this work, we find that interim semantic maps generated at intermediate layers can provide fine-grained prior information of object semantics distributions. Therefore, before stage $i$, our SPM incorporates the output feature $F_{i-1}^{out}$ and the interim semantic map $M_{i-1}$ of the previous stage to progressively learn reasonable visual prompts with a recurrent mechanism, since it is difficult to directly generate desirable prompts in some complex scenarios. After multiple iterations, we can obtain a refined semantic map $M_{i}$ and a suitable input $F_{i}^{in}$ for the stage $i$. This process can be formulated as:
\begin{equation}
  F_{i}^{in}, M_{i} = SPM(F_{i-1}^{out}, M_{i-1}, R), \label{eq:spm}
\end{equation}
where $R$ denotes the number of recurrent iterations. Notice that the initial semantic map $M_0$ is generated from the statistic category probability. More specifically, $M_0(x, y, c)$ at the position $(x, y)$ is set to the pixel probability of the $c$-th category on the downstream training set. Intuitively, our SPM works as a representation-level matched filter that can better match the input features with the pre-trained knowledge in the backbone. More details of the proposed SPM are described in Section \ref{sec:SPM}.

As shown in Figure \ref{fig:network-structure}, our SPM is hierarchically inserted between different stages to learn semantic-aware visual prompts for downstream tasks. Finally, the output feature $F_N^{out}$ of the $N$-th stage is fed into the head segmenter to generate the high-quality semantic map $M$.

\begin{figure}[t]
\centering
\includegraphics[width=0.992\columnwidth]{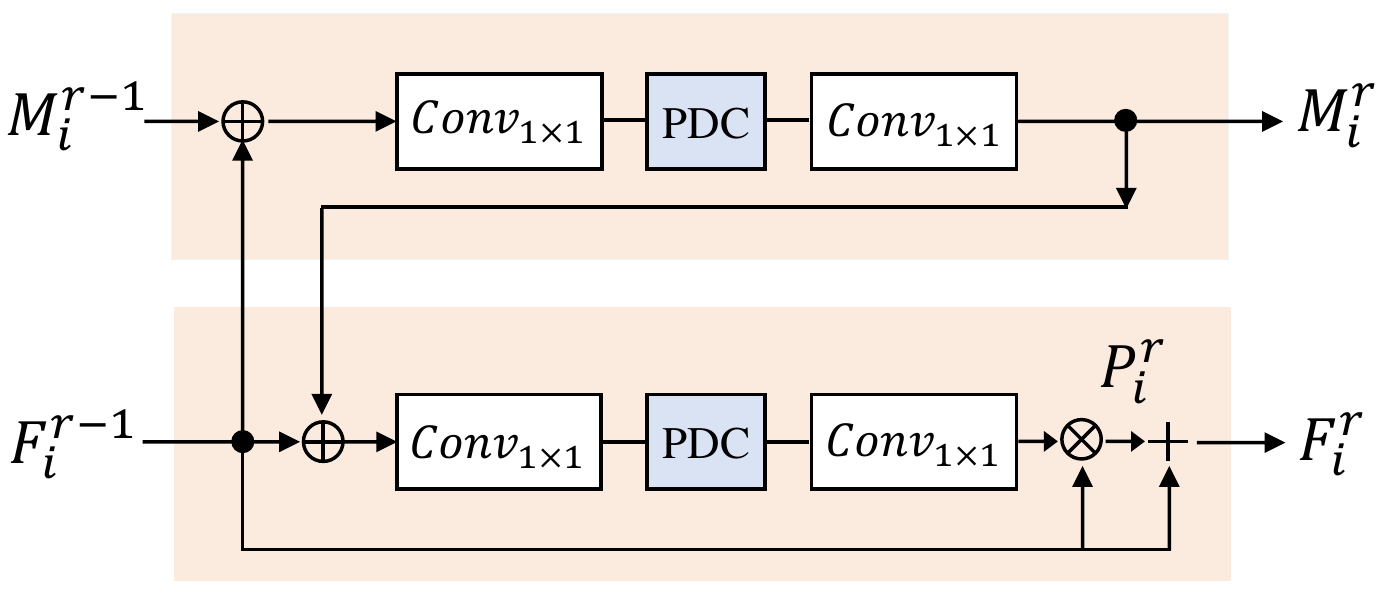}
\vspace{-3mm}
\caption{The architecture of Semantic-aware Prompt Matcher. $Conv_{1\times1}$ denotes a convolutional layer with a kernel size of $1\times1$ and PDC is our Pyramid Dilation Convolution in Figure \ref{fig:PDC}. $\oplus$ represents the feature concatenation operation, while $\otimes$ denotes the element-wise multiplication.}
\label{fig:SPM}
\end{figure}

\subsection{Semantic-aware Prompt Matcher} \label{sec:SPM}
In this subsection, we introduce the details of our SPM. The purpose of this module is to integrate rich information of interim semantic maps to learn reasonable visual prompts with a recurrent mechanism, so that the output feature of the previous stage can be transformed into a desirable input for the next stage. Specifically, before the stage $i$, our SPM first takes the feature $F_{i-1}^{out}$ and the semantic map $M_{i-1}$ to generate semantic-aware prompts. Similar to Recurrent Neural Networks \cite{lipton2015critical}, the prompted feature and the refined semantic map are fed back into SPM for recurrent prompt learning. Thus Eq.\ref{eq:spm} can be unfolded as follows:
%{\small
%\setlength\abovedisplayskip{4pt}
%\setlength\belowdisplayskip{4pt}
\begin{equation}
\begin{split}
  F_{i}^{0}, M_{i}^{0} &= F_{i-1}^{out}, M_{i-1}, \\
  F_{i}^{r}, M_{i}^{r} &= f(F_{i}^{r-1}, M_{i}^{r-1}, \theta) \text{~~~for}{~}r {~}\text{in}{~}\{1,...,R\}, \\
  F_{i}^{in}, M_{i}     &= F_{i}^{R}, M_{i}^{R}, \\
\end{split}
\end{equation}%
where $f(\cdot)$ denotes the SPM function with trainable parameters $\theta$. $F_{i}^{r}$ and $M_{i}^{r}$ are the prompted feature and the refined semantic map at the $r$-th iteration of the $i$-the SPM. After $R$ iterations, we can obtain the desirable input $F_{i}^{in}$ for the next stage to better energize the pre-trained knowledge. Notice that $\theta$ is shared for all iterations for parameter efficiency.

\begin{figure}[t]
\centering
\includegraphics[width=1\columnwidth]{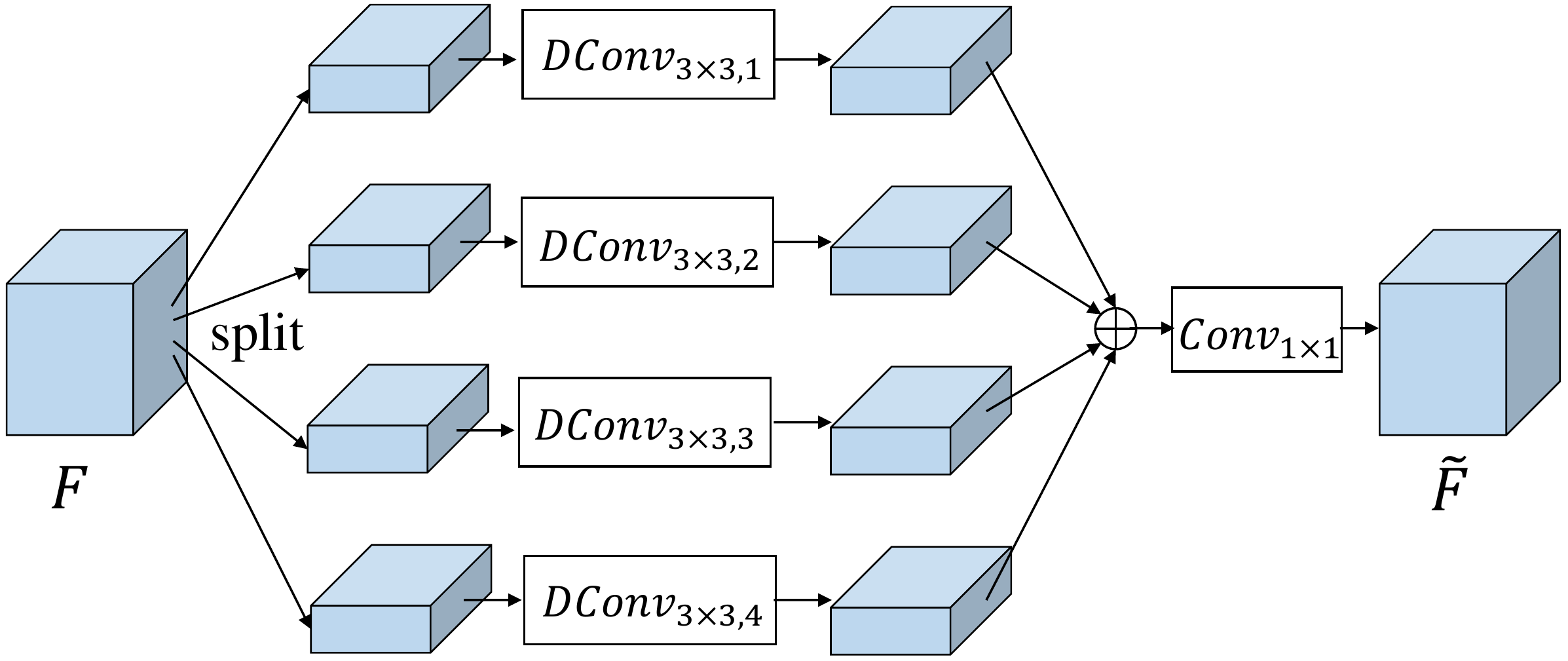}
\vspace{-3mm}
\caption{The architecture of Pyramid Dilation Convolution for long-range spatial context modeling. $DConv_{3\times3, d}$ denotes a dilated group convolutional layer with a kernel size of $3\times3$ and a dilated rate of $d$. The input feature $F$ and output feature $\tilde{F}$ have the same dimension.}
\label{fig:PDC}
\end{figure}

Figure \ref{fig:SPM} shows the visual prompt tuning for the $r$-th iteration of the $i$-the SPM. We can see that our SPM consists of two parallel branches to refine the interim semantic map and generate the prompted feature. As mentioned in previous works \cite{zhu2019asymmetric,he2019adaptive}, long-range spatial context is crucial for semantic segmentation. Thus we develop a Pyramid Dilation Convolution (PDC) that uses four dilated group convolutional layers to capture the multi-scale lang-range context. As shown in Figure \ref{fig:PDC}, the input feature of PDC is divided into four sub-features along the channel dimension. The $i$-th sub-feature is fed into the $i$-th convolutional layer with a kernel size of $3\times3$ and a dilated rate of $r$. The outputs of all dilated layers are concatenated and fused using a $1\times1$ convolutional layer to restore the original channel number. The proposed PDC is integrated into both branches to generate the long-range contextualized features. The details of these branches are described as follows.

{\bf{i) Interim Semantic Map Refinement: }} In this branch,  we utilize the current feature to enhance the semantic map generated from the previous feature. Specifically, we first feed the concatenation of $F_i^{r-1}$ and $M_i^{r-1}$ into a $1\times1$ group convolutional layer to generate a compact feature $F^{i,r}_M$, which has a low channel number $C$ for reducing the trainable parameters. We then apply the proposed PDC to obtain the long-range contextualized feature $\tilde{F}^{i,r}_M$, which is further fed into another $1\times1$ group convolutional layer to generate the refined semantic map $M_i^r$. This process can be formulated as:
%{\small
%\setlength\abovedisplayskip{7pt}
%\setlength\belowdisplayskip{7pt}
\begin{equation}
\begin{split}
    F^{i,r}_M &= Conv_{1\times1}(F_i^{r-1} \oplus M_i^{r-1}), \\
    \tilde{F}^{i,r}_M &= PDC(F^{i,r}_M), \\
    M_i^r &= Softmax\{Conv_{1\times1}(\tilde{F}^{i,r}_M)\},
\end{split}
\end{equation}%
where $\oplus$ denotes the feature concatenation operation and the $Softmax$ layer is used to normalize the predicted scores of semantic categories at each location.

{\bf{ii) Semantic-aware Prompt Generation: }}
We then incorporate the feature $F_i^{r-1}$ and the refined map $M_i^r$ to learn visual prompts progressively. Similar to the first branch, we feed $F_i^{r-1}$ and $M_i^r$ into a $1\times1$ convolutional layer and a PDC to obtain the features $F^{i,r}_P$ and $\tilde{F}^{i,r}_P$. Inspired by the attention mechanism, we utilize another $1\times1$ convolutional layer to generate a prompt weight $W^{i,k}_P$, which is further applied to multiply with $F_i^{k-1}$ to generate the visual prompt map $P_i^k$. Finally, $F_i^{k-1}$ and $P_i^k$ are added to obtain the new prompted feature $F_i^r$. This process can be formulated as:
{\small
\begin{equation}
\begin{split}
    F^{i,r}_P &= Conv_{1\times1}(F_i^{r-1} \oplus M_i^r), {~~} \tilde{F}^{i,r}_P = PDC(F^{i,r}_P), \\
    W^{i,r}_P &= Conv_{1\times1}(\tilde{F}^{i,r}_P),
    {~~~~~~~~~~~~~~~} P_{i}^r = F_i^{r-1} \otimes W^{i,r}_P, \\
    F_i^r &= F_i^{r-1} + P_{i}^r,
\end{split}
\end{equation}}%
where $\otimes$ denotes the element-wise multiplication operation. Thanks to the progressive visual prompts generated by $R$ iterations, we can effectively transform $F_{i-1}^{out}$ to an appropriate $F_{i}^{in}$ that can better stimulate the pre-trained knowledge in the $i$-th frozen stage of the backbone.

\subsection{Network Optimization}
%{\bf{Network Optimization: }}
During the training phase, the backbone network is frozen and we only update the parameters of our SPM and the head segmenter on downstream datasets. We utilize the Cross-Entropy (CE) loss function to optimize our network. The total loss is defined as follows:
{\small
\setlength\abovedisplayskip{2pt}
\setlength\belowdisplayskip{1pt}
\begin{equation}
  loss = CE(M, \hat{M}) + \sum_{i=1}^{N} \sum_{k=1}^{K} a_i * CE(M_i^k, \hat{M}),
  \label{eq:loss}
\end{equation}
}%
where $\hat{M}$ is the ground-truth semantic map and $M$ is our final semantic map predicted by the head segmenter. $a_i$ is the loss weight of those interim semantic maps generated by the $i$-th SPM.

\section{Experiment}
\subsection{Downstream Datasets}
\begin{figure*}[t]
  \begin{center}
     \includegraphics[width=1\linewidth]{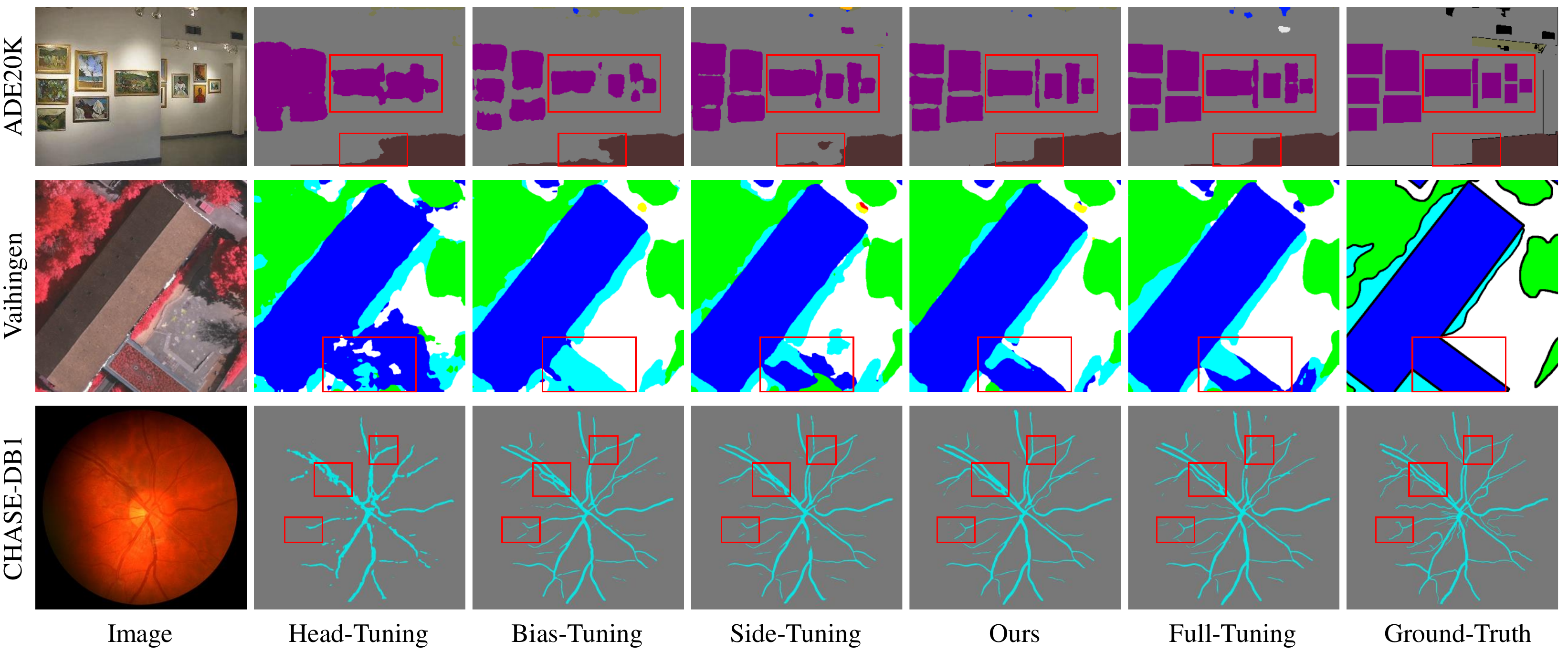}
  \vspace{-9mm}
  \end{center}
  \caption{Visualization of various samples of semantic segmentation and the results of different methods. We observe that our method can generate high-quality semantic maps for natural, satellite, and medical image segmentation, especially in the regions of red boxes.}
  \vspace{0mm}
  \label{fig:results}
\end{figure*}

In this work, we conduct extensive experiments on five semantic segmentation datasets of various scenarios. Some examples of these datasets are visualized in Figure \ref{fig:results} and their details are described as follows.

\textbf{ADE20K \cite{zhou2017scene}:} It is a large-scale scene parsing dataset with 150 types of objects and stuff. The public ADE20K consists of 20,210 images for training and 2,000 images for validation. These images have different resolutions and their objects suffer from great scale variations. This dataset is very challenging due to the high intra-class variances and low inter-class variances.

\textbf{Vaihingen \cite{vaihingen}:} It is a medium-scale dataset for satellite image segmentation. This dataset consists of six types of semantic object, e.g., buildings, streets, cars, etc. The training set contains 344 images, while the testing set has 398 images. All images has the same resolution $512\times512$.

\textbf{CHASE-DB1 \cite{fraz2012ensemble}, STARE \cite{hoover2000locating}: } They are small-scale medical image segmentation datasets that aim to segment retinal vessels from background. CHASE-DB1 consists of 28 retinal images of multiethnic school children. STARE contains 20 retinal images of people of various ages, 10 of which have pathology. We follow \cite{mmseg2020} to preprocess these datasets and partition them for training/testing.

\subsection{Comparison with the Stage of the Art}
In this subsection, we compare the proposed method with six basic and advanced methods, e.g., \textit{Full-Tuning} \cite{agrawal2014analyzing}, \textit{Learn-from-Scratch} \cite{zhu2019scratchdet}, \textit{Head-Tuning} \cite{donahue2014decaf}, \textit{Bias-Tuning} \cite{zaken2021bitfit}, \textit{Side-Tuning} \cite{zhang2020side} and \textit{Adapter} \cite{houlsby2019parameter}. The overall experimental settings are as follows:

{\bf{Backbone Network:}} Here all methods adopt the popular ResNet-101 \cite{he2016deep} network as backbone. Except \textit{Learn-from-Scratch}, the backbone parameters of all methods were pre-trained with an auxiliary task of image recognition on the ImageNet-1K dataset \cite{deng2009imagenet}. The experiments of the Transformer backbone can be referred to Section \ref{sec:ViT}.

{\bf{Head Segmenter:}} For ADE20K and Vaihingen, all methods use Pyramid Pooling Module \cite{zhao2017pyramid} as the head segmenter, since it has a great capacity to capture the scale variations of objects. On those medical datasets, the Progressive UPsampling head \cite{zheng2021rethinking} is used, because retinal vessels are very thin and we require high-resolution segmentation results for medical diagnosis. Notice that all parameters of these heads are randomly initialized.

\begin{table}[t]
  \caption{Performance and the number of trainable parameters of different methods on the ADE20K dataset.} %, when the backbone network is ResNet-101 pre-trained on ImageNet-1K.}
  \vspace{-2mm}
  \centering
  \resizebox{8.5cm}{!} {
    \begin{tabular}{c|c|c|c|c}
    \hline
    \multirow{2}{*}{Method} &
    \multicolumn{3}{c|}{Trainable Parameters (M)} &
    \multirow{2}{*}{mIoU $\uparrow$} \\
    \cline{2-4}
    & Backbone  & Prompt &  Head &  \\
    \hline\hline
    Full-Tuning \cite{agrawal2014analyzing}     & 42.41 & 0    & 25.54 & 43.96\\
    Learn-from-Scratch \cite{zhu2019scratchdet} & 42.41 & 0    & 25.54 & 34.84\\
    \hline
    Head-Tuning \cite{donahue2014decaf}         & 0     & 0    & 25.54 & 34.08\\
    Bias-Tuning \cite{zaken2021bitfit}          & 0.05  & 0    & 25.54 & 35.61\\
    Side-Tuning \cite{zhang2020side}            & 0     & 6.51 & 25.54 & 36.42\\
    Adapter \cite{houlsby2019parameter}         & 0     & 6.80 & 25.54 & 37.22\\
    \hline
    Ours                                        & 0     & 3.11 & 25.54 & {\bf{41.83}}\\
    \hline
    \end{tabular}
  }
  \label{tab:ADE20K}
\end{table}

{\bf{Trainable Parameters:}} During training, the head segmenter of each method is trainable. In addition, all methods except  \textit{Head-Tuning} will further optimize the whole/partial backbone network or introduce some prompt parameters. For instance, \textit{Full-Tuning} and \textit{Learn-from-Scratch} tune all backbone's parameters, while \textit{Bias-Tuning} tunes the bias of all layers in the backbone network. \textit{Side-Tuning} introduces a side-network between the input and output of each residual module, while \textit{Adapter} adds an adaptive-network on the output of the residual module. Both the side-network and adaptive-network are implemented by a $3\times3$ down-convolutional layer with 64 channels and 4 groups, and a $1\times1$ up-convolutional layer with the same number of channels as the output of the residual module. In our framework, the ResNet-101 backbone is divided into four stages, and the proposed SPM is inserted before each stage and the head segmenter to learn visual prompts hierarchically. The reduced channel $C$ in SPM is uniformly set to 256 and the group number of convolutional layers in PDC is set to 16.

\begin{table}[t]
  \caption{Performance and the number of trainable parameters of different methods on the Vaihingen dataset.} %when the backbone network is ResNet-101 pre-trained on ImageNet-1K.}
  \vspace{-2mm}
  \centering
  \resizebox{8.5cm}{!} {
    \begin{tabular}{c|c|c|c|c}
    \hline
    \multirow{2}{*}{Method} &
    \multicolumn{3}{c|}{Trainable Parameters (M)} &
    \multirow{2}{*}{mIoU $\uparrow$} \\
    \cline{2-4}
    & Backbone  & Prompt &  Head &  \\
    \hline\hline
    Full-Tuning \cite{agrawal2014analyzing}     & 42.41 & 0 & 25.43 & 73.96\\
    Learn-from-Scratch \cite{zhu2019scratchdet} & 42.41 & 0 & 25.43 & 68.85\\
    \hline
    Head-Tuning \cite{donahue2014decaf}         & 0     & 0 & 25.43 & 62.45\\
    Bias-Tuning \cite{zaken2021bitfit}          & 0.05  & 0 & 25.43 & 68.06\\
    Side-Tuning \cite{zhang2020side}            & 0     & 6.51 & 25.43 & 71.62\\
    Adapter \cite{houlsby2019parameter}         & 0     & 6.80 & 25.43 & 70.89\\
    \hline
     Ours                                       & 0     & 1.95 & 25.43 & {\bf{73.10}}\\
    \hline
    \end{tabular}
  }
  \label{tab:Vaihingen}
\end{table}

{\bf{Training Details:}} In this work, we apply MMSegmentation \cite{mmseg2020} to implement our experiments on four Nvidia GeForce 3090 GPUs. On the ADE20K and Vaihingen datasets, those compared methods are trained for 80K iterations with a batch size of 16, while our method is trained with a batch size of 12 due to the GPU memory limitation and the number of training iterations is set to 80K*16/12=106K. Therefore, all methods are optimized with the same epochs. For those medical datasets, all methods are trained for 8K iterations with a batch size of 4, due to the small amount of training data. The loss weight $a_i$ in Eq.\ref{eq:loss} is set to 0.05/R, 0.1/R, 0.2/R, 0.3/R, and 0.4/R for the five stages, respectively

\subsubsection{Large-scale Natural Image Segmentation}
Table \ref{tab:ADE20K} summarizes the performance of different methods on the ADE20K dataset. We can observe that \textit{Full-Tuning} achieves the best mIoU of 43.96\%, significantly outperforming \textit{Learn-from-Scratch} with the same number of trainable parameters, since the pre-trained backbone network contains rich prior knowledge. When the backbone is frozen, \textit{Head-Tuning} obtains a poor mIoU of 34.08\%. With around 6M extra parameters, \textit{Side-Tuning} and \textit{Adapter} boost the mIoU to 36.42\% and 37.22\% respectively. In contrast, our method achieves a promising mIoU of $41.83\%$, using only $3.11$M extra parameters to learn visual prompts between different frozen stages. In summary, the proposed method is better than all other parameter-efficient methods and also comparable to \textit{Full-Tuning}. Moreover, Figure \ref{fig:results} also shows that our method can generate high-quality semantic maps for unconstrained natural scenarios.

\begin{table}[t]
  \caption{The foreground Dice Similariy Coefficient (Dice) of one-shot medical image semantic segmentation.} %when the backbone network is ResNet-101 pre-trained on ImageNet-1K.}
  \vspace{-2mm}
  \centering
  \resizebox{8.5cm}{!} {
    \begin{tabular}{c|c|c|c|c|c}
    \hline
    \multirow{2}{*}{Method} &
    \multicolumn{3}{c|}{Trainable Parameters (M)} &
    \multicolumn{2}{c}{Dice $\uparrow$} \\
    \cline{2-6}
    & Backbone  & Prompt &  Head & CHASE-DB1 & STARE\\
    \hline\hline
    Full-Tuning \cite{agrawal2014analyzing}      & 42.41 & 0    & 8.26 & 76.07$\pm$0.57 & 75.90$\pm$1.98\\
    Learn from Scratch \cite{zhu2019scratchdet}  & 42.41 & 0    & 8.26 & 73.20$\pm$1.11 & 73.01$\pm$0.84\\
    \hline
    Head-Tuning \cite{donahue2014decaf}          & 0     & 0    & 8.26 & 59.35$\pm$0.55 & 54.14$\pm$1.82\\
    Bias-Tuning \cite{zaken2021bitfit}           & 0.05  & 0    & 8.26 & 75.83$\pm$0.68 & 74.70$\pm$1.11\\
    Side-Tuning \cite{zhang2020side}             & 0     & 6.51 & 8.26 & 76.22$\pm$0.51 & 75.45$\pm$1.06\\
    Adapter \cite{houlsby2019parameter}          & 0     & 6.80 & 8.26 & 76.07$\pm$0.63 & 75.13$\pm$0.93\\
    \hline
    Ours                                         & 0     & 1.94 & 8.26 & {\bf{77.08$\pm$0.68}} & {\bf{76.46$\pm$1.16}}\\
    \hline
    \end{tabular}
  }
  \vspace{0mm}
  \label{tab:medical_seg}
\end{table}

\subsubsection{Medium-scale Satellite Image Segmentation}
Table \ref{tab:Vaihingen} shows the performance of all methods on the Vaihingen dataset. We can observe that the compared results on this dataset are consistent with those on the ADE20K dataset. More specifically, \textit{Head-Tuning} performs poorly when the backbone network is frozen. Without exception, \textit{Side-Tuning} and \textit{Adapter} are better than \textit{Bias-Tuning}, but their results are still not satisfactory. In contrast, our method achieves a competitive mIoU of $73.10\%$, when using only $1.95$M extra parameters for prompt learning. There is a small gap between our method and \textit{Full-Tuning} in terms of performance. What's more, our method only needs to store a small number of parameters. This experiment shows the great potential of the proposed method for medium-scale satellite image segmentation.

\subsubsection{One-shot Medical Image Segmentation}
We further apply the proposed method to address one-shot medical semantic segmentation, since it is difficult and expensive to obtain massive medical images with pixel-wise annotations. Here we conduct five one-shot experiments. In each experiment, we randomly select a sample to train models and then validate them on the whole testing set. Table \ref{tab:medical_seg} shows the mean results and variances of different methods on two retinal segmentation datasets. We notice that those parameter-efficient fine-tuning methods are comparable to or even better than \textit{Full-Tuning} under the one-shot setting, which is consistent with the findings \cite{brown2020language,zeng2021pangu} in the NLP field. In particular, our method can achieve the best results on both datasets by learning visual prompts effectively with only $1.95$M extra trainable parameters. The above experiments show the generalization of our method for semantic segmentation of various scenarios.

\begin{table}[t]
  \caption{Performance of different prompted stages and their number of prompt parameters (M) on three representative datasets. The recurrent number $R$ of SPM is set to 1 in this experiment.}
  \vspace{-2mm}
  \centering
  \resizebox{8.7cm}{!} {
    \begin{tabular}{c|c|c|c|c|c|c}
    \hline
    \multirow{2}{*}{Stages} & \multicolumn{2}{c|}{ADE20K} & \multicolumn{2}{c|}{Vaihingen} & \multicolumn{2}{c}{CHASE-DB1} \\
    \cline{2-7}
     & \#Params & mIoU &  \#Params & mIoU & \#Params & Dice \\
    \hline\hline
    1 {~~}  & 0.48 & 34.43      & 0.40 & 69.64      & 0.40 & 74.32$\pm$0.79\\
    1-2     & 1.07 & 35.11      & 0.87 & 71.26      & 0.87 & 76.30$\pm$0.56\\
    1-3     & 1.78 & 36.39      & 1.48 & 71.76      & 1.47 & 76.67$\pm$0.74\\
    1-4     & 2.36 & 38.54      & 1.95 & {\bf72.60} & 1.94 & {\bf{77.08$\pm$0.68}}\\
    1-5     & 3.11 & {\bf40.01} & 2.55 & 72.17      & 2.54 & 76.40$\pm$0.75\\
    \hline
    \end{tabular}
  }
  \vspace{0mm}
  \label{tab:Prompted_Stages}
\end{table}

\subsection{Ablation Studies} \label{sec:ablation}
In this subsection, we validate the impact of each module of our method on three datasets. The hyper-parameters validated on CHASE-DB1 are directly used with STARE.

{\textbf{Effects of Different Prompted Stages:}}
We first explore the effects of inserting the proposed SPM into varied stages of the backbone. As mentioned above, ResNet-101 consists of 4 stages and we treated the head segmenter as the fifth stage. As shown in Table \ref{tab:Prompted_Stages}, our performance gradually increases as the number of prompted stages increases. More specifically, our method achieves a promising mIoU with five prompted stages on ADE20K. For Vaihingen and CHASE-DB1, the best number of prompted stages is 4, and more prompted stages no longer result in obvious improvements. Thus, we can conclude that more prompted stages are required on large-scale datasets and fewer prompted stages on small/medium-scale datasets.

{\textbf{Effects of Different Recurrent Iterations of SPM:}}
We then explore the recurrent prompt learning mechanism of our SPM. As shown in Table \ref{tab:iteration}, on the ADE20K dataset, our method can obtain better results as the recurrent number $R$ increases, and it achieves the best mIoU 41.83\% when $R$ is set to 3. This is because that downstream segmentation tasks on ADE20K have a large domain gap with the pre-trained knowledge from ImageNet, and our SPM needs more recurrent iterations to learn effective prompts progressively. On the Vaihingen dataset, our SPM requires two recurrent iterations to obtain the best results, as the object variances of Vaihingen are smaller than that of ADE20K. On the CHASE-DB1 dataset, with only one iteration, our SPM has a sufficient capacity to generate effective visual prompts and achieve superior performance. Thus we can draw a conclusion that large-scale challenging datasets require more recurrent iterations of prompt learning, while small/medium-scale datasets require fewer iterations.

\begin{table}[t]
  \caption{Performance of different recurrent numbers of SPM on three representative datasets.}
  \vspace{-2mm}
  \centering
  \resizebox{8.2cm}{!} {
    \begin{tabular}{c|c|c|c}
    \hline
    \#Recurrent & ADE20K & Vaihingen & CHASE-DB1 \\
    \hline\hline
    1    & 40.01        & 72.60        & {\bf{77.08$\pm$0.68}}\\
    2    & 41.08        & {\bf{73.10}} & 76.24$\pm$0.79\\
    3    & {\bf{41.83}} & 72.24        & -\\
    \hline
    \end{tabular}
  }
  \vspace{0mm}
  \label{tab:iteration}
\end{table}

\begin{table}[t]
  \caption{Performance of our method with/without semantic-aware prompt learning (SPL) and long-range spatial context modeling (LSCM) on three representative datasets.}
  \vspace{-2mm}
  \centering
  \resizebox{8.2cm}{!} {
    \begin{tabular}{c|c|c|c}
    \hline
    Method & ADE20K & Vaihingen & CHASE-DB1 \\
    \hline\hline
    w/o SPL   & 39.99 & 71.03 & 75.88$\pm$0.56\\
    w/o LSCM  & 40.06 & 72.26 & 76.45$\pm$0.78\\
    Ours      & 41.83 & 73.10 & 77.08$\pm$0.68\\
    \hline
    \end{tabular}
  }
  \label{tab:Semantic-Aware}
\end{table}

{\textbf{Effects of Semantic-aware Prompt Learning: }}
In our SPM, interim semantic maps are incorporated to learn visual prompts. Here we ablate this model design by removing the guided semantic information. Specifically, we set all those weights $a_i$ in Eq.\ref{eq:loss} to 0, making our network learn prompts in a black-box mapping manner. Table \ref{tab:Semantic-Aware} (Row 1) shows that removing the semantic-aware information will lead to inferior results on all datasets. For instance, mIoU drops from 41.83\% to 39.99\% on ADE20K, while Dice decreases from 77.08\% to 75.88\% on CHASE-DB1. These experiments illustrate that the prior information of interim semantic maps is beneficial for effective prompt learning.

{\textbf{Effects of Long-range Spatial Context Modeling: }}
We further explore the effectiveness of long-range spatial context modeling. Here we implement a variant of SPM that does not explicitly capture long-range context by setting the dilated rate of all convolutional layers in PDC to 1. As shown in Table \ref{tab:Semantic-Aware} (Row 2), the mIoU of this variant drops to 40.06\% for on ADE20K and 72.26\% on Vaihingen, while its Dice decreases to 76.45\% on CHASE-DB1. These experiments indicate that the long-range spatial context is meaningful for semantic segmentation prompt learning.

\begin{figure}[t]
\centering
\includegraphics[width=1\columnwidth]{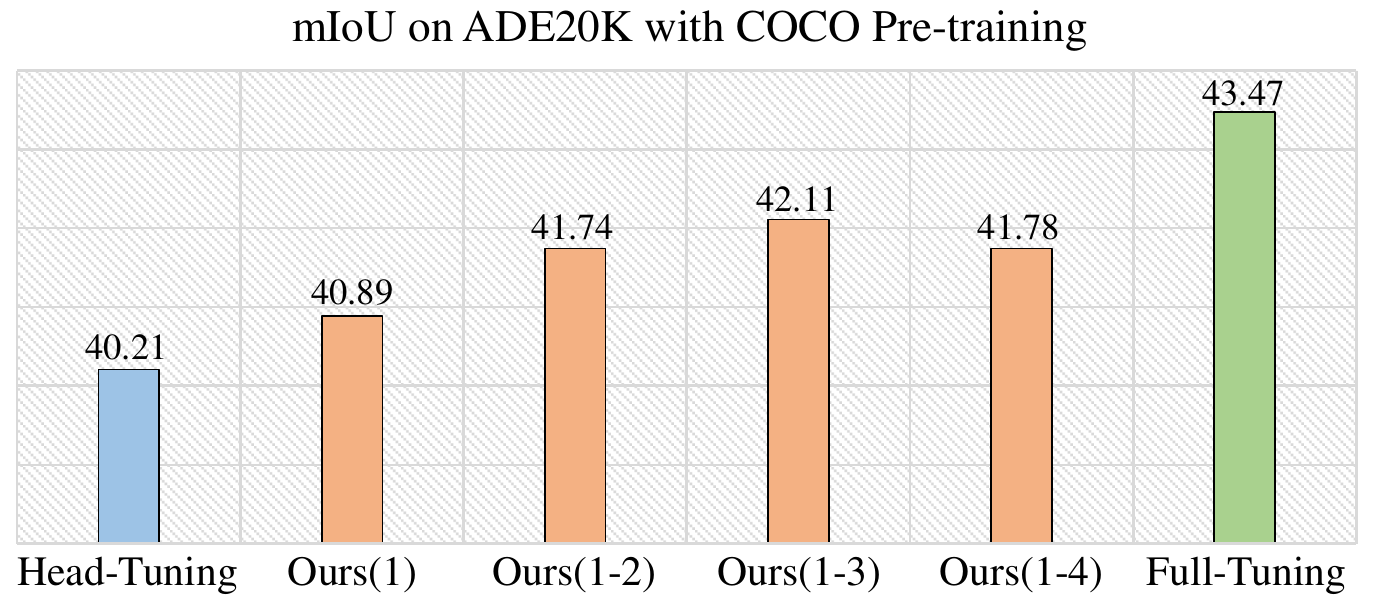}
\vspace{-6mm}
\caption{Performance on the ADE20K dataset when the backbone network was pre-trained on the COCO dataset. Ours($i$-$j$) denotes a variant of our method that inserts SPM before stages $i$-$j$.}
\label{fig:coco-pretraining}
\end{figure}

{\bf{Effects of Different Pre-training Datasets: }}
Finally, we explore the influences of pre-training data from different source domains. Besides the backbone pre-trained on ImageNet, we reimplement our experiments using the ResNet-101 backbone pre-trained on the COCO-Stuff-164K dataset \cite{caesar2018coco}, another large-scale semantic segmentation benchmark. Figure \ref{fig:coco-pretraining} summarizes the results of different methods based on the COCO pre-trained backbone. Compared with the results of ImageNet pre-training in Table \ref{tab:Prompted_Stages}, we observe that our method can achieve a better mIoU of $42.11\%$ by using fewer prompted stages. This phenomenon indicates that it is beneficial to choose models pre-trained on a proper source domain. Nevertheless, many downstream tasks do not have large-scale datasets in practice.  Therefore it makes sense to adopt the foundation models pre-trained on ImageNet, since they have good generalization.

\subsection{Apply to Transformer} \label{sec:ViT}
As mentioned above, our method is also generic to the Transformer architecture. Thus in this subsection, we apply the proposed method to fine-tune the large-scale Vision-Transformer (ViT-L) \cite{zheng2021rethinking}, which consists of 24 transformer layers. More specifically, we divide these layers into $N$ stages evenly and perform visual prompt learning before each stage. Meanwhile, we also explore the recurrent mechanism of SPM based on the backbone ViT-L. Following \cite{jia2022visual}, the learning rate multiplier of prompts is set to 10, so that the head segmenter and prompts share the same learning rate. Table \ref{tab:Ablation_ViT} summarizes the performance of five variants of our method, which are trained for 160K iterations with a batch size of 8. We can observe that our method achieves a competitive mIoU 45.05\% on the ADE20K dataset, when ViT-L is divided into three stages and each SPM performs twice prompt learning. Hence we adopt this setting for ViT-L prompt tuning, which only introduces 1.76M extra prompt parameters.

Based on the ViT-L backbone, we further compare the proposed method with six stage-of-the-art approaches on ADE20K. As shown in Table \ref{tab:ADE20K_Vit}, our method is much better than VPT \cite{jia2022visual} that learns ten prompt tokens of each transformer layer. Moreover, with fewer prompt parameters, our method also outperforms the recent AdaptFormer \cite{chen2022adaptformer} and the ensemble VPT+Bias-Tuning, which learns 100 prompt tokens for each layer and fine-tunes backbone's bias simultaneously. This is mainly attributed to that our lightweight SPM can generate effective visual prompts using rich information of interim semantic maps. These experiments show the promising potential of our method for large-scale Transformer fine-tuning.

\begin{table}[t]
\caption{Ablation Studies of different stage numbers and SPM recurrent numbers of our method on ADK20K when the backbone is the ViT-L pre-trained on ImageNet-21K \cite{ridnik2021imagenet}.}
\label{tab:Ablation_ViT}
\vspace{-2mm}
\centering
\begin{tabular}{c|c|c|c}
    \hline
    \#Stage & \#Recurrent & \#Prompt (M) & mIoU $\uparrow$ \\
    \hline\hline
    1 & 1 & 0.59 & 42.47\\
    2 & 1 & 1.17 & 44.28\\
    3 & 1 & 1.76 & 44.72\\
    \hline
    3 & 2 & 1.76 & {\bf{45.05}}\\
    3 & 3 & 1.76 & 44.60\\
    \hline
\end{tabular}
\end{table}

\begin{table}[t]
\caption{Performance of different methods on ADE20K when the backbone is ViT-L pre-trained on ImageNet-21K. Those methods with * are trained with a batch size of 16 and their results are quoted from \cite{jia2022visual}, while other methods are implemented by us with a batch size of 8 due to the limited GPU resources. All methods are optimized for 160K iterations.}
\label{tab:ADE20K_Vit}
\vspace{-2mm}
\centering
\resizebox{8.35cm}{!} {
\begin{tabular}{c|c|c|c|c}
    \hline
    \multirow{2}{*}{Method} &
    \multicolumn{3}{c|}{Trainable Parameters (M)} &
    \multirow{2}{*}{mIoU $\uparrow$} \\
    \cline{2-4}
    & Backbone  & Prompt &  Head &  \\
    \hline\hline
    Full-Tuning \cite{agrawal2014analyzing} & 304.15 & 0 & 13.14 & 47.53\\
    \hline
    Head-Tuning \cite{donahue2014decaf}     & 0    & 0    & 13.14 & 37.77\\
    VPT* \cite{jia2022visual}               & 0    & 0.25 & 13.14 & 42.11\\
    Bias-Tuning* \cite{zaken2021bitfit}     & 0.28 & 0    & 13.14 & 43.40\\
    AdaptFormer \cite{chen2022adaptformer}  & 0    & 3.17 & 13.14 & 44.00\\
    VPT+Bias-Tuning*                        & 0.28 & 2.50 & 13.14 & 44.04\\
    \hline
    Ours  & 0 & 1.76 & 13.14 & {\bf{45.05}}\\
    \hline
\end{tabular}
}
\end{table}

\begin{table*}[t]
    \centering
    \caption{The top-1 accuracy of different fine-tuning methods for image recognition, when the backbone network is ViT-B pre-trained on ImageNet-21K. We also report the tunable parameter percentage and accuracy gap of various methods relative to \textit{Full-Tuning} in brackets.}
    \vspace{0mm}
    \begin{tabular}{l | c  |c| c |c }
    \hline
    Method     &  Avg Params (M)     & {CIFAR-100} & {SVHN} & {Food-101} \\
    \hline
    Full-Tuning \cite{agrawal2014analyzing} & {86.04}~(100\%) &{85.90} & {97.67} & {90.09}  \\
    \hline
    Head-Tuning \cite{donahue2014decaf}     & ~~0.07~(0.08\%) & 69.83~\bbb{-16.07}    & 66.91~\bbb{-30.76}    & 69.74~\bbb{-20.35}  \\
    VPT \cite{jia2022visual}                & ~~0.08~(0.09\%) & 82.44~\bbb{-3.46}{~~} & 94.02~\bbb{-3.65}{~~} & 82.98~\bbb{-7.11}{~~} \\
    AdaptFormer \cite{chen2022adaptformer}  & ~~1.26~(1.46\%) & 85.90~\aaa{+0.00}{~~}  & 96.89~\bbb{-0.78}{~~} & 87.61~\bbb{-2.48}{~~} \\
    \hline
    Ours                                    & ~~1.00~(1.16\%) & 86.62~\aaa{+0.72}{~~} & 97.49~\bbb{-0.18}{~~} & 88.63~\bbb{-1.46}{~~} \\
    \hline
    \end{tabular}
    \label{tab:recognition}
\end{table*}

\section{Conclusion}
In this paper, we propose a universal Stage-wise Prompt-Matched Framework to fine-tune foundation models of CNN/Transformer to handle various downstream tasks. A lightweight Semantic-aware Prompt Matcher is introduced to learn effective visual prompts progressively between different stages of the frozen backbone network, thereby better stimulating pre-trained knowledge and promoting downstream representation learning. We conduct extensive experiments on four datasets of semantic segmentation to verify the performance effectiveness and parameter efficiency of our method. In the future, we would apply the proposed method to more downstream tasks.

\begin{figure}[t]
\vspace{0mm}
\centering
\includegraphics[width=8.5cm]{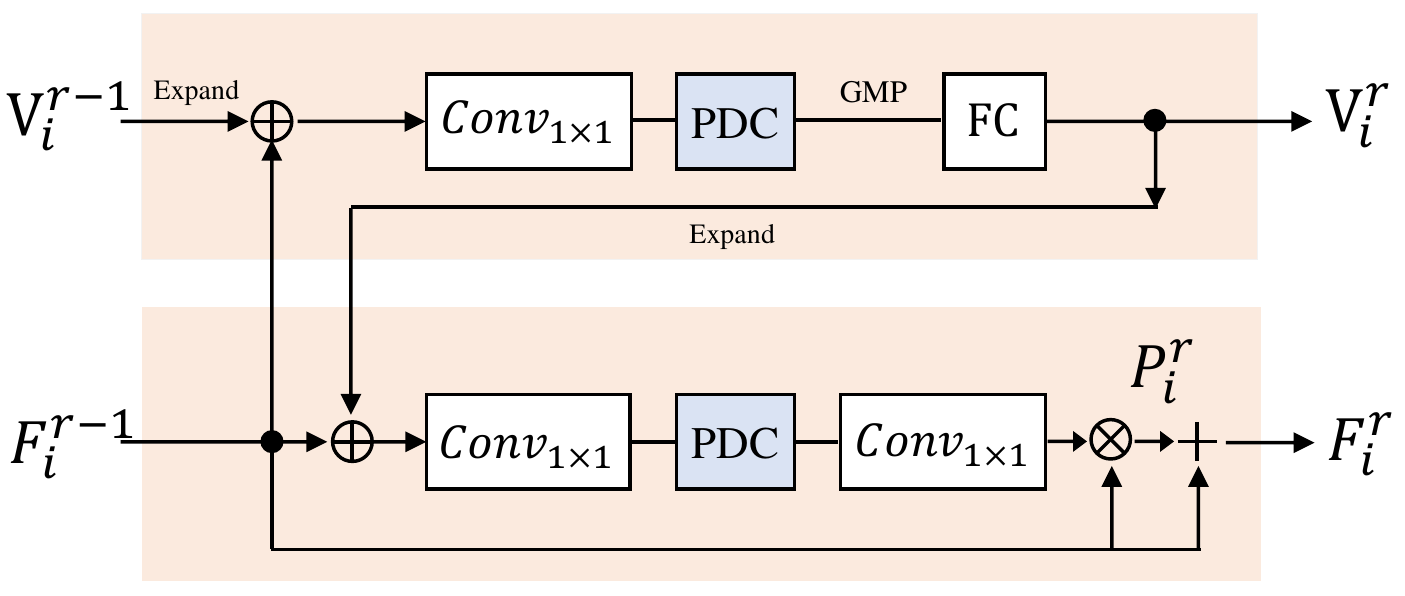}
\vspace{-3mm}
\caption{The architecture of Semantic-aware Prompt Matcher for image recognition. Here the interim category probability vector $V_i^r$ is used to learn semantic-aware visual prompts. GMP is a global max-pooling layer, while FC represents a fully-connected layer. ''Expand`` means that $V_i^{r-1}$ and $V_i^{r}$ are expanded (via copying) to have the same resolution as $F_i^{r-1}$ before feature concatenation.}
\label{fig:SPM_class}
\end{figure}

\section*{Appendix: Apply to Image Recognition}
It is worth noting that our method is general to various visual tasks. In this supplementary material, we apply the proposed Stage-wise
Prompt-Matched Framework to effectively fine-tune pre-trained foundation models for image recognition. Following the previous work \cite{chen2022adaptformer}, we take ViT-B \cite{zheng2021rethinking} as the backbone network, which consists of 12 transformer layers and was pre-trained on ImageNet-21K with the self-supervised method MAE \cite{he2022masked}. We freeze the parameters of the whole backbone and insert four SPM before layers 1,5,9,12 to hierarchically learn visual prompts. More specifically, as shown in Figure \ref{fig:SPM_class}, the interim semantic map $M_i^{r-1}$ in our SPM is replaced by the interim category probability vector $V_i^{r-1} \in \mathbb{R}^{C_d}$, where ${C_d}$ denotes the category number of the downstream image recognition dataset. In the first branch, the expanded $V_i^{r-1}$ and feature $F_i^{r-1}$ are concatenated and fed into a $1{\times}1$ convolutional layer, a PDC module, a global max-pooling layer, and a fully-connected layer to generate the refined probability vector $V_i^{r}$. The second branch is the same as in the main text. Noth that the initial semantic map $P_0$ is set to the statistic category probability of the training set of the downstream dataset. Here the channel number of our SPM is set to 96 for image recognition. We completely follow the hyper-parameter setting of AdaptFormer \cite{chen2022adaptformer} to conduct experiments on three representative datasets, including CIFAR-100 \cite{krizhevsky2009learning}, SVHN \cite{goodfellow2013multi} and Food-101 \cite{bossard2014food}.

The top-1 accuracy of different fine-tuning methods for image recognition is summarized in Table \ref{tab:recognition}. We can observe that the proposed method achieves promising performance on all benchmarks, outperforming the state-of-the-art VPT \cite{jia2022visual} and AdaptFormer \cite{chen2022adaptformer} consistently. In particular, our method is also better than \textit{Full-Tuning} \cite{agrawal2014analyzing} on CIFAR-100, and is comparable on SVHN and Food-101. Besides the performance superiority, our method also has great advantages in terms of parameter efficiency. Specifically, our method only need to optimize 1.06M parameter for CIFAR-100/Food-101 and 0.88M parameter for SVHN. On average, the amount of our tunable parameters of our method is 1\% of that of \textit{Full-Tuning}. These experiments demonstrate the great potential of our method for image recognition fine-tuning.

%%%%%%%%% REFERENCES
{\small
\bibliographystyle{ieee_fullname}
\bibliography{egbib}
}

\end{document}